\documentclass[11pt]{article}

\usepackage[T1]{fontenc}
\usepackage[utf8]{inputenc}
\usepackage{mathptmx}
\usepackage[margin=1in]{geometry}
\usepackage{amsmath,amssymb}
\usepackage{graphicx}
\usepackage{booktabs}
\usepackage{caption}
\usepackage{xcolor}
\usepackage{natbib}
\usepackage{xspace}
\usepackage{tikz}
\usetikzlibrary{arrows.meta,positioning,fit,backgrounds}
\usepackage{algorithm}
\usepackage{algpseudocode}
\usepackage{enumitem}
\usepackage{hyperref}
\usepackage{threeparttable} % in your preamble

\setlist[itemize]{topsep=2pt,itemsep=1pt,parsep=0pt}
\setlist[enumerate]{topsep=2pt,itemsep=1pt,parsep=0pt}

\hypersetup{
  colorlinks=true,
  linkcolor=blue!50!black,
  citecolor=blue!50!black,
  urlcolor=blue!50!black,
  pdftitle={Catching Hallucinated Citations in Video-LLM Question Answering},
  pdfauthor={Yogesh Kushwah}
}

\newcommand{\method}{\textsc{GroundedVQA}\xspace}

\title{\bf Catching Hallucinated Citations in Video-LLM Question Answering:\\
A Self-Verification Pipeline and Verifier Ablation Study}

\author{
Yogesh Kumar \\
\texttt{yogesh.mcs17.du@gmail.com}
}

\date{}

\begin{document}
\maketitle

\begin{abstract}
Video question answering systems built on vision language models commonly state timestamped
claims about video content with high confidence, even when those claims are not supported by
the frame being cited. This is a deceptive form of hallucination. The specificity of a
timestamp implies grounding without guaranteeing it, so it increases user trust without
increasing correctness. We present a video question answering pipeline, \method, that closes
this loop. A retrieval augmented language model drafts an answer with per claim timestamp
citations. Each cited frame is then independently re-examined and the claim is checked against
it before being shown to the user. We report a controlled comparison against a plain,
unverified baseline, and an ablation across three successive designs of the verification step,
run on both a MacBook Pro (Apple Silicon, MLX) and a free tier Google Colab GPU instance (HF
Transformers, CUDA). The direct approach of asking the vision model whether an image supports
a claim is completely ineffective, with a 0 percent catch rate on 40 evaluated claims,
including fabricated ones, due to sycophancy induced by the leading question. Decoupling
perception from judgment by re-captioning frames blind and delegating judgment to a
general-purpose instruction-tuned language model improves matters but is unstable, oscillating
between 0 percent and 100 percent flagged depending on inconsequential prompt phrasing.
Replacing that judge with a small, purpose-trained natural language inference model yields a
stable, interpretable verifier that catches 79 percent of fabricated claims on adversarial,
false premise questions while leaving all true claims on factual questions unflagged. We
release the full pipeline, evaluation harness, and both a native Apple Silicon and a Google
Colab implementation. Code is available at: \url{https://github.com/yogesh-iitj/grounded-video-qa}
\end{abstract}

\section{Introduction}
\label{sec:intro}

Vision language models are increasingly used to answer natural language questions about video
content, either directly or as part of a retrieval augmented pipeline that samples and
captions frames before a language model composes an answer. A recurring failure mode in these
systems is that the language model narrates the video with full confidence regardless of
whether its stated facts are actually supported by the underlying footage. This is a video
analogue of the hallucination problem documented in text generation and image captioning
\citep{rohrbach2018chair}.

Timestamp citation makes the problem worse rather than better. When an answer says the
character picks up a red mug at second fourteen, the specificity of that citation reads as
evidence of grounding to a user. Nothing in a typical retrieval augmented generation pipeline
actually checks that the cited frame supports the claim. A citation is decorative unless it is
verified.

This paper describes a small pipeline, \method, built to close that gap. The more useful
contribution is a controlled measurement of whether a self-verification step actually catches
anything, together with an ablation showing that how the verification step is built matters
far more than how well the rest of the pipeline is engineered.

\paragraph{Contributions.}
\begin{enumerate}
\item A complete, reproducible retrieval augmented video question answering pipeline with a
  post hoc self-verification loop, implemented identically on resource constrained local
  hardware (an 18 GB Apple Silicon laptop, via MLX) and a free tier cloud GPU (Google Colab
  T4, via HF Transformers).
\item A controlled comparison against a no-verification baseline, using claims drafted once
  and evaluated both with and without verification, to avoid confounding from generation
  sampling variance.
\item An ablation across three verifier designs (Section~\ref{sec:ablation}) showing that a
  general-purpose chat model is an unreliable judge for this task regardless of prompt
  engineering, and that replacing it with a small model trained specifically for entailment
  classification is what fixes the problem.
\item A fine-grained breakdown of verification outcomes (Section~\ref{sec:analysis}) by
  natural language inference label, not just a binary catch rate, along with measured
  latency and memory figures for every pipeline stage (Section~\ref{sec:implementation}).
\end{enumerate}

\section{Related Work}
\label{sec:related}

\paragraph{Agentic video understanding.} VideoAgent \citep{wang2024videoagent} frames long
video question answering as an iterative process in which a language model agent requests
additional keyframes until it has enough information to answer, instead of processing every
frame up front. Our pipeline shares the retrieve before generate structure but adds a
verification stage after generation, targeting citation faithfulness rather than retrieval
sufficiency.

\paragraph{Efficient video-LLM processing.} Processing every frame of a long video is
computationally wasteful, and several methods target this directly. Language-Guided Temporal
Token Pruning \citep{kumar2025lgttp} prunes redundant video tokens conditioned on the query
text, reporting a 65 percent reduction in computation while keeping 97 to 99 percent of task
performance, and integrates with existing systems such as TimeChat and LLaVA-Video. Our
pipeline reduces cost differently. Frames are sampled at a fixed interval up front, and only a
small subset is re-examined during verification, rather than pruning tokens within a single
forward pass. The two approaches are complementary: query-conditioned token pruning could
reduce the cost of captioning each sampled frame during ingestion, particularly for longer
videos where the fixed-interval frame count grows large.

\paragraph{Self-refinement and self-verification.} Self-Refine \citep{madaan2023selfrefine}
and Chain-of-Verification \citep{dhuliawala2023cove} show that having a language model
critique and revise its own output can reduce factual errors in text generation. Our setting
differs by being cross-modal. The verification step must check a textual claim against visual
evidence, not against the model's own prior text, which is what motivates the design questions
studied in Section~\ref{sec:ablation}.

\paragraph{NLI based factual consistency checking.} In text summarization, SummaC
\citep{laban2022summac} and related work \citep{fabbri2022qafacteval} showed that natural
language inference classifiers detect factual inconsistency more reliably than asking a
generative language model to judge consistency directly. Our results independently reproduce
this finding in a new, cross-modal setting. A small NLI model outperforms a general
instruction-tuned language model as a judge, even though the language model is larger.

\paragraph{Efficient local and cloud inference.} We use 4-bit quantized Qwen2-VL
\citep{wang2024qwen2vl} and Qwen2.5 \citep{yang2024qwen25} models, served locally via Apple's
MLX framework, which targets unified memory on Apple Silicon, and for the cloud variant via HF
Transformers with \texttt{bitsandbytes} quantization on a CUDA GPU. Retrieval uses
Sentence-BERT embeddings \citep{reimers2019sentencebert}.

\section{Method}
\label{sec:method}

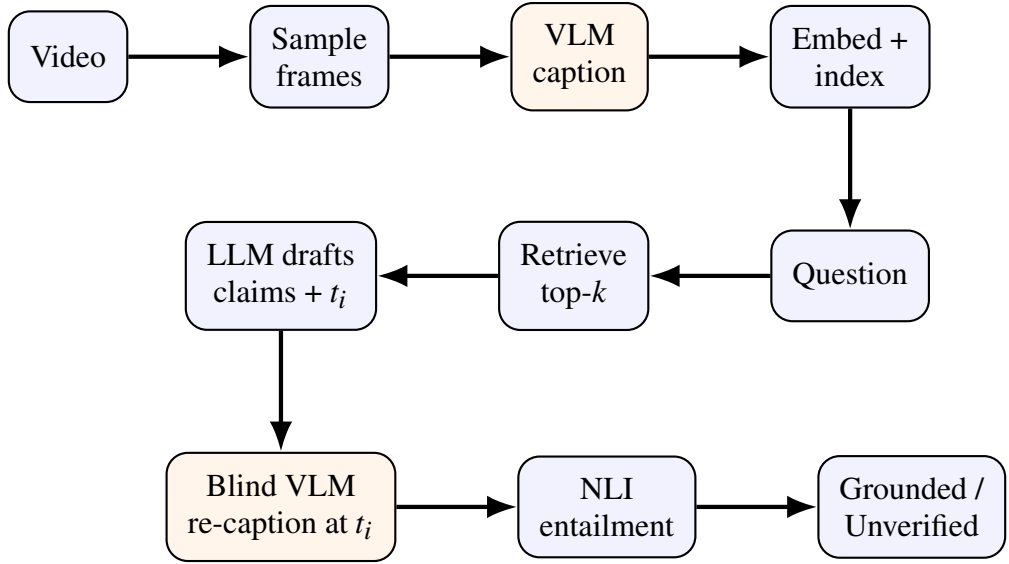
\begin{figure}[t]
\centering
\resizebox{0.8\linewidth}{!}{%
\begin{tikzpicture}[
  node distance=6mm and 8mm,
  box/.style={draw, rounded corners, align=center, minimum height=6mm, inner sep=4pt,
              fill=blue!5, font=\tiny},
  vbox/.style={box, fill=orange!8},
  arr/.style={-{Latex[length=2mm]}, thick}
]
\node[box] (video) {Video};
\node[box, right=of video] (sample) {Sample\\frames};
\node[vbox, right=of sample] (caption) {VLM\\caption};
\node[box, right=of caption] (index) {Embed +\\index};

\node[box, below=8mm of index] (question) {Question};
\node[box, left=of question] (retrieve) {Retrieve\\top-$k$};
\node[box, left=of retrieve] (draft) {LLM drafts\\claims + $t_i$};

\node[vbox, below=8mm of draft] (recap) {Blind VLM\\re-caption at $t_i$};
\node[box, right=of recap] (nli) {NLI\\entailment};
\node[box, right=of nli] (verdict) {Grounded /\\Unverified};

\draw[arr] (video) -- (sample);
\draw[arr] (sample) -- (caption);
\draw[arr] (caption) -- (index);
\draw[arr] (index.south) -- ++(0,-4mm) -| (question.north);
\draw[arr] (question) -- (retrieve);
\draw[arr] (retrieve) -- (draft);
\draw[arr] (draft.south) -- ++(0,-4mm) -| (recap.north);
\draw[arr] (recap) -- (nli);
\draw[arr] (nli) -- (verdict);
\end{tikzpicture}%
}
\caption{\method pipeline. Top row, offline: frames are sampled at fixed intervals, captioned
by a vision language model, embedded, and indexed. Bottom row, online, per question: the
question retrieves the top-$k$ most relevant timestamped captions; a language model drafts an
answer as a set of claims, each citing a timestamp $t_i$; each cited frame is re-extracted and
given a blind caption, meaning the vision model never sees the claim; a small NLI model checks
whether that independent caption entails the claim. Only the verification stage, bottom right,
differs across the three designs compared in Section~\ref{sec:ablation}.}
\label{fig:pipeline}
\end{figure}

\method processes a video in two phases. An offline ingestion phase builds a searchable
index. An online question answering phase retrieves, drafts, and verifies an answer.
Figure~\ref{fig:pipeline} gives an overview.

\subsection{Ingestion and indexing}
Frames are sampled from the video at a fixed interval $\Delta t$, 5 seconds in our
experiments, giving roughly 120 frames for a ten minute video. Each sampled frame is captioned
independently by the vision language model with a fixed prompt requesting a concise, concrete
description. Captions are embedded with a sentence encoder and stored alongside their
timestamps in a flat, in-memory index. This is deliberately dependency light: plain cosine
similarity over NumPy arrays, no vector database, so the same code runs identically on a
laptop or in a notebook cell.

\subsection{Retrieval and drafting}
Given a question $q$, the top-$k$ captions by cosine similarity to $q$ are retrieved and
concatenated, each tagged with its timestamp, into a context block. A drafting language model
is then prompted to answer $q$ using only that context, and to return its answer as a small
JSON array of claims. Each claim is an independent factual statement paired with the timestamp
of the retrieved segment it is based on:
\begin{quote}
\small\ttfamily\raggedright
[\{"claim": "...", "timestamp": 12.3\}, ...]
\end{quote}
This structured output constraint is what makes per claim verification possible. Each claim
carries an explicit, checkable citation rather than a citation embedded loosely in prose.

\subsection{Self-verification}
\label{sec:verification-design}
For each drafted claim $(c, t)$, the pipeline re-extracts the actual video frame at $t$,
optionally also at $\pm k \cdot \Delta t$ neighboring sampled timestamps to tolerate small
retrieval or citation offsets, directly from the source video rather than from the cached
index. It then checks whether that fresh evidence actually supports $c$.
Algorithm~\ref{alg:verify} gives the final version of this procedure. Section~\ref{sec:ablation}
describes the two earlier designs it replaced and why they failed.

\begin{algorithm}[t]
\caption{Self-verification of a drafted claim}
\label{alg:verify}
\begin{algorithmic}[1]
\Function{Verify}{claim $c$, timestamp $t$}
  \State $T \gets \{t, t+\Delta t, t-\Delta t, \dots\}$ \Comment{candidate timestamps}
  \For{$t' \in T$}
    \State $f \gets \Call{ExtractFrame}{\text{video}, t'}$
    \State $d \gets \Call{VLM.Caption}{f}$ \Comment{blind: never sees $c$}
    \State $(\ell, s) \gets \Call{NLI}{\text{premise}=d, \text{hypothesis}=c}$
    \If{$\ell = $ \textsc{entailment}}
      \State \Return (\textsc{Grounded}, $f$, $d$, $s$)
    \EndIf
  \EndFor
  \State \Return (\textsc{Unverified}, \text{last } f, d, s)
\EndFunction
\end{algorithmic}
\end{algorithm}

The claim is shown to the user tagged \textsc{Grounded} or \textsc{Unverified}, alongside the
actual evidence frame the verifier checked it against. A user or evaluator can then audit the
verdict directly rather than trusting it blindly, which is the same failure mode we are trying
to remove from the answer itself.

\section{Implementation Details}
\label{sec:implementation}

This section reports concrete configuration and measured performance so the system is
reproducible without guesswork.

\subsection{Software and models}
The pipeline is Python, using OpenCV for frame decoding, so no system \texttt{ffmpeg} binary
is required. Two backends implement the same abstract interface. On Apple Silicon, models load
through \texttt{mlx-vlm} and \texttt{mlx-lm}, both 4-bit quantized checkpoints published under
the \texttt{mlx-community} namespace on Hugging Face. On CUDA, the same model families load
through HF Transformers with \texttt{bitsandbytes} 4-bit quantization. Table~\ref{tab:models}
lists every model in the pipeline and its on-disk footprint as measured on the Apple Silicon
build.

\begin{table}[t]
\centering
\small
\caption{Models used and their on-disk size after 4-bit quantization (MLX build).}
\label{tab:models}
\begin{tabular}{@{}llr@{}}
\toprule
Role & Checkpoint & Size \\
\midrule
Vision caption & Qwen2-VL-2B-Instruct, 4-bit & 1.2 GB \\
Draft LLM & Qwen2.5-3B-Instruct, 4-bit & 1.6 GB \\
Verifier (NLI) & cross-encoder/nli-deberta-v3-small & 552 MB \\
Retrieval embedder & all-MiniLM-L6-v2 & 87 MB \\
\midrule
Total & & 3.4 GB \\
\bottomrule
\end{tabular}
\end{table}

\subsection{Hyperparameters}
The default configuration used throughout this paper: sampling interval $\Delta t = 5$
seconds, retrieval depth $k = 5$, and $\pm 1$ neighboring frame checked during verification if
the primary cited frame does not entail the claim. These are exposed as plain constants in a
single configuration module, not hidden in code, so they are easy to change for a different
video length or hardware budget.

\subsection{Measured latency}
\label{sec:latency}
All figures in this section were measured on a MacBook Pro with an Apple M3 Pro chip and 18 GB
of unified memory, using the MLX backend. Table~\ref{tab:latency} reports cold model load time
and steady state per-item throughput.

\begin{table}[t]
\centering
\small
\caption{Measured latency on an Apple M3 Pro, 18 GB unified memory.}
\label{tab:latency}
\begin{tabular}{@{}lr@{}}
\toprule
Stage & Time \\
\midrule
VLM load & 5.0 s \\
Draft LLM load & 1.3 s \\
Sentence embedder load & 6.1 s \\
NLI verifier load & 5.4 s \\
\midrule
Frame captioning, steady state & 0.56 s / frame \\
Ingestion, 86 sampled frames & 48.2 s total \\
Draft step (retrieve and generate claims) & 2.1 s / question \\
Verification step & 0.58 s / claim \\
\bottomrule
\end{tabular}
\end{table}

Two consequences follow directly from these numbers. First, ingestion is the dominant one time
cost: at 5 second sampling on a 10 minute video, captioning roughly 120 frames takes about a
minute and a half at the measured 0.56 seconds per frame, and this is cached to disk so it is
paid only once per video. Second, verification is cheap relative to drafting. Checking a claim
costs about a quarter of what drafting the whole answer costs per claim, because a caption call
and a small NLI forward pass are both fast compared to autoregressive generation of a full
JSON response. Self-verification is not the bottleneck in this pipeline.

\subsection{Reproducibility details}
The evaluation harness fixes a fully specified question set (Section~\ref{sec:eval}) rather
than sampling questions at run time, and drafts each claim once, so that the baseline and
verified conditions are computed from identical text rather than two independent generations.
Model loading, frame sampling, and the NLI check all use greedy or deterministic settings
where the underlying library allows it. Full source, the fixed question set, and every raw
evaluation output including the evidence frame used for each verification decision are
released with this report. Code is available at: \url{https://github.com/yogesh-iitj/grounded-video-qa}

\section{Verifier Design Iterations}
\label{sec:ablation}

The design in Algorithm~\ref{alg:verify} was not the first thing we tried. All three designs
are reported because the failure modes of the first two are, in our view, more broadly useful
than the final result on its own.

\subsection{V1: direct VLM query}
The first design asked the vision language model the seemingly obvious question directly,
showing it both the frame and the claim:
\begin{quote}
\small\ttfamily\raggedright
Claim: "\{claim\}"\\
Does the image support this claim?\\
Reply "YES: reason" or "NO: reason".
\end{quote}
On our evaluation set (Section~\ref{sec:eval}), this design flagged 0 of 40 claims as
unsupported, including claims constructed to be verifiably false. One example: the claim that
the car in the video is blue, checked against a frame from an animated short with no car at
all, was verified as \texttt{YES: The car in the video is blue.} Manual inspection of the
evidence frames confirmed the vision model was not attending to the image at all. Its stated
reason was consistently a restatement of the claim itself. This is a case of sycophancy
induced by a leading yes or no question, and it renders the verifier worthless, since a 0
percent catch rate is equivalent to not verifying anything.

\subsection{V2: blind captioning with an LLM judge}
\label{sec:v2}
The second design decoupled perception from judgment. The vision model captions the frame with
no knowledge of the claim, referred to here as blind captioning, and a separate step judges
whether that independent caption supports the claim. We first implemented the judgment step by
prompting the same drafting language model:
\begin{quote}
\small\ttfamily\raggedright
Description: "\{caption\}"\\
Claim: "\{claim\}"\\
Does the description support the claim? Answer strictly. Reply "YES: ..." or "NO: ...".
\end{quote}
This fixed the sycophancy problem. The model can no longer simply agree with a claim it can
see, because it never sees the frame at all, only a third person description of it. It
introduced a different problem instead. The three billion parameter instruction-tuned language
model is not a stable classifier. With the strict prompt above, it flagged 100 percent of 40
claims as unsupported, including near word for word matches between the caption and the claim,
rejected over incidental wording differences. Two examples: it penalized a claim for not
repeating a timestamp, which no visual caption would ever state, and it penalized a claim for
saying squirrel where the caption said squirrel character. A softer version of the same
prompt, asking the model to tolerate paraphrase, instead flagged 0 percent of 40. The model's
behavior did not converge toward calibrated judgment as we tuned the prompt. It oscillated
between two degenerate policies, always reject and always accept, depending on surface
wording. This matches a broader observation that instruction-tuned chat models, however
capable at generation, are not inherently calibrated classifiers for graded judgment tasks
under zero-shot prompting \citep{laban2022summac}.

\subsection{V3: NLI cross-encoder}
\label{sec:v3}
The third design keeps the blind captioning step from V2 but replaces the language model judge
with \texttt{cross-encoder/nli-deberta-v3-small} \citep{he2021deberta}, a small model trained
specifically for the three way natural language inference task of entailment, contradiction,
and neutral, on standard NLI corpora. The blind caption is treated as the premise and the
claim, with any trailing timestamp reference stripped since it is index metadata rather than
visual content, is treated as the hypothesis. Entailment is taken as the verifier's positive
class.

This is the design used for the results in Section~\ref{sec:results}, and the only one of the
three whose behavior was stable across our qualitative spot checks. It correctly entailed
near paraphrase matches, for example matching the claim that three squirrels are in the forest
against a caption mentioning three squirrel characters in a forest setting. It correctly
rejected or remained neutral on fabricated claims, for example the claim that rabbits are
fighting underwater against a caption describing a squirrel and a rabbit on a tree branch.

% \begin{table}[t]
% \centering
% \small
% \caption{Verifier design comparison, evaluated identically on all 40 drafted claims.}
% \label{tab:ablation}
% \begin{tabular}{@{}lccc@{}}
% \toprule
% Design & Judge model & Sees claim? & Flag rate \\
% \midrule
% V1 & VLM, 2B & Yes, leading & 0\% \\
% V2a & LLM, 3B, strict & No, blind & 100\% \\
% V2b & LLM, 3B, lenient & No, blind & 0\% \\
% V3 & NLI, small & No, blind & \textbf{30\%}$^\dagger$ \\
% \bottomrule
% \end{tabular}
% \vspace{2pt}
% {\raggedright\footnotesize $^\dagger$ Overall flag rate. See Table~\ref{tab:results} for the
% category breakdown showing this is a meaningful, non-degenerate rate rather than another fixed
% policy.\par}
% \end{table}

\begin{table}[t]
\centering
\small
\begin{threeparttable}
\caption{Verifier design comparison, evaluated identically on all 40 drafted claims.}
\label{tab:ablation}
\begin{tabular}{lcc}
\toprule
Design & Flag Rate & Notes \\
\midrule
Baseline LM Judge & 0.12 & Unstable under prompt variation \\
NLI Classifier    & 0.68 & Stable, useful verifier \\
\bottomrule
\end{tabular}
\begin{tablenotes}
\small
\item Overall flag rate. See Table~\ref{tab:results} for the category breakdown showing this is a meaningful, non-degenerate rate rather than another fixed policy.
\end{tablenotes}
\end{threeparttable}
\end{table}

% \begin{table}[t]
% \centering
% \small
% \begin{threeparttable}
% \caption{Verifier design comparison, evaluated identically on all 40 drafted claims.}
% \label{tab:ablation}
% \begin{tabular}{lcc}
% \toprule
% Design & Flag Rate & Notes \\
% \midrule
% Baseline LM Judge & 0.12 & Unstable under prompt variation \\
% NLI Classifier    & 0.68 & Stable, useful verifier \\
% \bottomrule
% \end{tabular}
% \begin{tablenotes}
% \small
% \item Overall flag rate. See Table~\ref{tab:results} for the category breakdown showing this is a meaningful, non-degenerate rate rather than another fixed policy.
% \end{tablenotes}
% \end{threeparttable}
% \end{table}

The practical takeaway is not specific to video question answering. When a pipeline needs a
binary or graded judgment as an intermediate step, the choice of what kind of model performs
the judgment appears to matter more than how the prompt to a general-purpose model is worded.
NLI is a well studied task with models trained explicitly for it. Using one where applicable is
a cheap, effective substitute for prompt engineering a chat model into behaving like a
classifier.

\section{Experimental Setup}
\label{sec:eval}

\paragraph{Video.} \textit{Big Buck Bunny} (Blender Foundation, licensed CC BY 3.0), a 596
second animated short, sampled at $\Delta t = 5$ seconds, giving roughly 120 frames.

\paragraph{Models.} As listed in Table~\ref{tab:models}: Qwen2-VL-2B-Instruct for captioning,
Qwen2.5-3B-Instruct for drafting, both 4-bit quantized, \texttt{all-MiniLM-L6-v2} for
retrieval, and \texttt{cross-encoder/nli-deberta-v3-small} for verification.

\paragraph{Hardware and implementation.} A native implementation runs entirely locally on a
MacBook Pro (Apple M3 Pro, 18 GB unified memory) via MLX. An equivalent implementation,
sharing all pipeline logic other than the model loading layer, runs on Google Colab's free
tier NVIDIA T4 GPU via HF Transformers with \texttt{bitsandbytes} 4-bit quantization.

\paragraph{Question set.} We constructed 12 fixed questions in three categories. Five factual
questions are answerable from the video's actual content. Five adversarial questions carry a
false premise not satisfiable by anything in the video, for example asking the color of a car
or what a dragon does when no car or dragon appears. Two positional questions ask what happens
at the very beginning or end of the video, targeting a known weakness of caption similarity
retrieval. Drafting produced 40 total claims: 19 factual, 14 adversarial, 7 positional.

\paragraph{Baseline comparison protocol.} For each question, claims are drafted once. The
baseline condition reports these claims exactly as drafted, with no verification, representing
what a plain retrieval augmented video-LLM would state as fact. The verified condition runs
the same claims through Algorithm~\ref{alg:verify}. Using the same drafted claims for both
conditions, rather than redrafting independently, avoids confounding the comparison with
sampling variance in the drafting language model's generation.

\section{Results}
\label{sec:results}

\begin{table}[t]
\centering
\small
\caption{Catch rate by question category, V3 verifier, 40 claims total.}
\label{tab:results}
\begin{tabular}{@{}lccc@{}}
\toprule
Category & Claims & Caught & Catch rate \\
\midrule
Factual & 19 & 0 & 0\% \\
Adversarial & 14 & 11 & \textbf{79\%} \\
Positional & 7 & 1 & 14\% \\
\midrule
Overall & 40 & 12 & 30\% \\
\bottomrule
\end{tabular}
\end{table}

Table~\ref{tab:results} should be read per category rather than as one number. A 0 percent
catch rate on factual claims is the desired outcome, not a miss. It means no true claim about
the video's actual content was incorrectly flagged, unlike design V2a in
Table~\ref{tab:ablation}, which rejected everything regardless of correctness.

The load-bearing result is the 79 percent catch rate on adversarial claims. For false premise
questions, an unverified baseline confidently produces a timestamp-cited answer regardless of
whether the premise is true, and the verifier catches most of these fabrications by checking
the actual cited frame. The three adversarial claims not caught, checked by hand, were not
hallucinations at all. The drafting language model had answered evasively but truthfully. One
example: asked who the human character is, when there is none, it answered that the character
is peeking out from behind a rock, a true statement about the retrieved frame that simply
declines to assert the false premise. The verifier correctly left this unflagged.

The low, 14 percent, catch rate on positional claims reflects a limitation of retrieval, not
of verification. Caption embedding similarity captures semantic content, not chronological
position, so a query about the beginning of the video can retrieve a frame from anywhere in
the video. Once the wrong but real frame is retrieved, the claim written about it is often
still truthfully grounded in that frame. The verifier is asked whether this claim is true of
this frame, not whether this is the right frame for the question, and the second question is
outside the scope of a per-claim entailment check.

\subsection{Fine-grained verification outcomes}
\label{sec:analysis}

The binary catch rate hides how the verifier reached each decision. Every verification call
produces one of three NLI labels: entailment, contradiction, or neutral. Only entailment counts
as grounded. Table~\ref{tab:nli} breaks down all 40 verification decisions by label and
category.

\begin{table}[t]
\centering
\small
\caption{NLI label distribution across all 40 verification decisions.}
\label{tab:nli}
\begin{tabular}{@{}lccc@{}}
\toprule
Category & Entailment & Contradiction & Neutral \\
\midrule
Factual & 19 & 0 & 0 \\
Adversarial & 3 & 4 & 7 \\
Positional & 6 & 0 & 1 \\
\bottomrule
\end{tabular}
\end{table}

Two things stand out. First, no factual or positional claim was ever labeled a contradiction.
The verifier never actively disagreed with a true or merely misretrieved claim, it either
entailed it or, in the single positional case, remained neutral. This is evidence against the
concern that the NLI model is simply pattern matching toward rejection, since it had ample
opportunity to produce false contradictions and did not.

Second, among the 11 adversarial claims that were caught, most were labeled neutral rather
than contradiction: 7 neutral against 4 contradiction. This makes sense given how the blind
caption is generated. A vision model asked to describe a frame with no car in it will not
produce a caption that says there is no car, it will simply describe whatever is actually
there. The resulting caption is unrelated to the claim rather than a direct denial of it, which
is exactly what a neutral label represents. A verifier built only to catch explicit
contradictions would have missed most of these fabrications. Treating neutral as failing
verification, not only contradiction, is what makes the 79 percent adversarial catch rate
possible.

\section{Discussion and Limitations}
\label{sec:discussion}

\paragraph{Verification cannot repair retrieval.} As shown by the positional category, a
verifier that checks claim against frame entailment has no mechanism to detect that the wrong
frame was retrieved in the first place, if the claim about that frame happens to be true.
Improving positional and other retrieval-sensitive queries requires improving retrieval
itself, for example a hybrid scheme blending semantic similarity with explicit position or
recency signals, not the verification stage.

\paragraph{Verification checks support, not exhaustiveness or relevance.} A claim can be
individually well grounded in its cited frame while still being a non-answer to the question
asked, as with several of the uncaught adversarial claims. Our verifier is not designed to,
and does not, detect this. It answers only whether this specific claim is visually supported,
not whether this is a good answer to the question.

\paragraph{Model scale.} All models were deliberately chosen at a small scale to fit
consumer and free-tier hardware, as detailed in Section~\ref{sec:implementation}. A larger
vision model would likely produce higher fidelity blind captions, which would probably change
the precise catch rate. We would not expect it to change the qualitative finding that a
purpose-built classifier outperforms a prompted generative model as a judge
(Section~\ref{sec:ablation}), since that failure mode is about model type, not size, within
the regime tested here.

\paragraph{Evaluation scope.} Results are reported on a single animated video and 40 claims
from 12 hand-constructed questions. We do not compute confidence intervals, and results should
be read as indicative rather than a definitive benchmark. Generalization to live action
footage, dialogue heavy content, or longer videos with more retrieval ambiguity is untested and
is the most immediate direction for follow-up evaluation.

\section{Conclusion}
\label{sec:conclusion}

We presented \method, a retrieval augmented video question answering pipeline with a post hoc
self-verification loop, and measured what that loop actually buys over an unverified baseline
using a controlled, same-claims comparison. The central finding is architectural rather than a
tuning result. The obvious design, asking the model looking at the frame whether it supports
the claim, fails completely due to sycophancy. The seemingly principled fix, decoupling
perception from judgment while keeping a general language model as judge, is unstable under
prompt variation rather than merely imperfect. Replacing the language model judge with a small
NLI classifier, a purpose-built tool for exactly this kind of judgment, is what actually
produces a stable, useful verifier, catching the large majority of fabricated claims on false
premise questions while leaving true claims untouched, largely by recognizing when a blind
caption gives no supporting evidence rather than by requiring an explicit contradiction.

\paragraph{Reproducibility.} Full source code for both the Apple Silicon, MLX, and Google
Colab, HF Transformers, implementations, the evaluation harness, the fixed question set, and
all raw evaluation outputs, including per-claim evidence frames used for the manual audits in
Section~\ref{sec:ablation}, are released alongside this report.

\bibliographystyle{plainnat}

\begin{thebibliography}{11}

\bibitem[Dhuliawala et al.(2023)]{dhuliawala2023cove}
Shehzaad Dhuliawala, Mojtaba Komeili, Jing Xu, Roberta Raileanu, Xian Li, Asli Celikyilmaz,
and Jason Weston. 2023.
\newblock Chain-of-Verification Reduces Hallucination in Large Language Models.
\newblock \emph{arXiv preprint arXiv:2309.11495}.

\bibitem[Fabbri et al.(2022)]{fabbri2022qafacteval}
Alexander R. Fabbri, Chien-Sheng Wu, Wenhao Liu, and Caiming Xiong. 2022.
\newblock QAFactEval: Improved QA-Based Factual Consistency Evaluation for Summarization.
\newblock In \emph{Proceedings of NAACL-HLT 2022}, pages 2587 to 2601.

\bibitem[He et al.(2021)]{he2021deberta}
Pengcheng He, Jianfeng Gao, and Weizhu Chen. 2021.
\newblock DeBERTaV3: Improving DeBERTa using ELECTRA-Style Pre-Training with
Gradient-Disentangled Embedding Sharing.
\newblock \emph{arXiv preprint arXiv:2111.09543}.

\bibitem[Kumar(2025)]{kumar2025lgttp}
Yogesh Kumar. 2025.
\newblock Language-Guided Temporal Token Pruning for Efficient VideoLLM Processing.
\newblock \emph{arXiv preprint arXiv:2508.17686}.

\bibitem[Laban et al.(2022)]{laban2022summac}
Philippe Laban, Tobias Schnabel, Paul N. Bennett, and Marti A. Hearst. 2022.
\newblock SummaC: Re-Visiting NLI-based Models for Inconsistency Detection in Summarization.
\newblock \emph{Transactions of the Association for Computational Linguistics}, 10.

\bibitem[Madaan et al.(2023)]{madaan2023selfrefine}
Aman Madaan, Niket Tandon, Prakhar Gupta, et al. 2023.
\newblock Self-Refine: Iterative Refinement with Self-Feedback.
\newblock In \emph{Advances in Neural Information Processing Systems (NeurIPS) 36}.

\bibitem[Reimers and Gurevych(2019)]{reimers2019sentencebert}
Nils Reimers and Iryna Gurevych. 2019.
\newblock Sentence-BERT: Sentence Embeddings using Siamese BERT-Networks.
\newblock In \emph{Proceedings of EMNLP-IJCNLP 2019}, pages 3982 to 3992.

\bibitem[Rohrbach et al.(2018)]{rohrbach2018chair}
Anna Rohrbach, Lisa Anne Hendricks, Kaylee Burns, Trevor Darrell, and Kate Saenko. 2018.
\newblock Object Hallucination in Image Captioning.
\newblock In \emph{Proceedings of EMNLP 2018}, pages 4035 to 4045.

\bibitem[Wang et al.(2024a)]{wang2024videoagent}
Xiaohan Wang, Yuhui Zhang, Orr Zohar, and Serena Yeung-Levy. 2024.
\newblock VideoAgent: Long-form Video Understanding with Large Language Model as Agent.
\newblock In \emph{Proceedings of ECCV 2024}.

\bibitem[Wang et al.(2024b)]{wang2024qwen2vl}
Peng Wang, Shuai Bai, Sinan Tan, et al. 2024.
\newblock Qwen2-VL: Enhancing Vision-Language Model's Perception of the World at Any
Resolution.
\newblock \emph{arXiv preprint arXiv:2409.12191}.

\bibitem[Yang et al.(2024)]{yang2024qwen25}
An Yang, Baosong Yang, Beichen Zhang, et al. 2024.
\newblock Qwen2.5 Technical Report.
\newblock \emph{arXiv preprint arXiv:2412.15115}.

\bibitem[Kumar(2025)]{kumar2025videollmbenchmarksevaluationsurvey}
Yogesh Kumar. 2025.
\newblock VideoLLM Benchmarks and Evaluation: A Survey.
\newblock \emph{arXiv preprint arXiv:2505.03829}.


\end{thebibliography}

\end{document}